\documentclass{article}
\usepackage{spconf,amsmath,graphicx,subcaption}
\usepackage{makecell}
\usepackage{adjustbox}
\usepackage{algorithm}
\usepackage{algorithmic}
\title{ Meta-cognition-based Simple and Effective Approach to Object Detection} 
\name{Sannidhi P Kumar$^{\star}$ \qquad Chandan Gautam$^{\star}$ \qquad Suresh Sundaram$^{\star}$\thanks{We would like to thank WIRIN for funding this work.}}
\address{$^{\star}$ WIRIN Autonomous Systems and Robotics Lab  \\
Department of Aerospace Engineering\\ Indian Institute of Science, Bangalore, Karnataka}
\begin{document}
%\ninept
%
\maketitle
\begin{abstract}
Recently, many researchers have attempted to improve deep learning-based object detection models, both in terms of accuracy and operational speeds. However, frequently, there is a trade-off between speed and accuracy of such models, which encumbers their use in practical applications such as autonomous navigation. In this paper, we explore a meta-cognitive learning strategy for object detection to improve generalization ability while at the same time maintaining detection speed. The meta-cognitive method selectively samples the object instances in the training dataset to reduce overfitting. We use YOLO v3 Tiny as a base model for the work and evaluate the performance using the MS COCO dataset. The experimental results indicate an improvement in absolute precision of $2.6\%$ (minimum), and $4.4\%$ (maximum), with no overhead to inference time.  
\end{abstract}
\section{INTRODUCTION}
\label{sec:intro}

Object detection is one of the fundamental computer vision tasks used in a plethora of applications such as autonomous navigation, surveillance and security, and facial detection. Such widespread use necessitates the need to build accurate real-time object detection networks. Object detection combines the tasks of image classification and object localization. Thus, an object detector's task is to return the bounding box coordinates of the objects of interest in an image and assign them class labels. Modern deep learning-based object detectors comprise two parts, a backbone pre-trained on ImageNet and a detection head. Several backbones, such as \cite{VGG,ResNet,ResNext,DenseNet,MobileNet} have been proposed. Detection heads are classified into two types, two-stage object detectors, and one-stage object detectors. Two-stage detectors, such as Region-based Convolutional Neural Networks (R-CNN) \cite{RCNN}, Fast R-CNN \cite{FastRCNN}, Faster R-CNN \cite{FasterRCNN}, and Region-based Fully Convolutional Networks \cite{RFCN}, generate region proposals in the first stage. In the second stage, feature extraction from these region proposals is followed by object classification and bounding box regression. These detectors have achieved high accuracy rates but low operational speeds. To address this issue, single-stage detectors have been proposed. These detectors skip the region proposal stage and make bounding box predictions directly from the input image. As a result, these models are faster but not as accurate as two-stage detectors. Among the single-stage detectors, Single Shot MultiBox Detector (SSD) \cite{SSD}, the You Only Look Once (YOLO) series of networks, and RetinaNet \cite{RetinaNet} are the most representative. SSD directly produces class predictions and adjustments to default bounding boxes of different aspect ratios and scales for each location in a feature map. YOLO \cite{YOLO} frames object detection as a regression problem wherein network evaluation is performed just once to predict the bounding boxes and class probabilities. YOLO v2 \cite{YOLOv2} improves the Average Precision (AP) of YOLO significantly by introducing batch normalization for convolutional layers, a high-resolution classifier, and dimension clusters. YOLO v3 \cite{YOLOv3} proposes a Darknet-53 architecture that further improves upon the performance of YOLO v2. YOLO v4  \cite{YOLOv4}, being the latest addition to the family, introduces new features such as weighted residual connections, mosaic data augmentation, cross mini-batch normalization, cross-stage partial connections, self adversarial training, and Complete Intersection over Union (CIoU) loss. 

Variants of the above described YOLO series have been released for constrained environments that meet real-time requirements. One such model is YOLO v3 Tiny, which clocks at $220$ FPS, but gives an AP of $33.1\%$.  The objective is to improve accuracy while preserving the detection speed. It is recently shown in cognitive psychology that human meta-cognition learning principles help achieve better generalization ability.  Over the last decade, these principles have been used across various disciplines such as disease classification, science education, human emotion recognition, and search algorithms for optimization problems. Various researchers have amalgamated the concept of meta-cognition with neural networks in the past \cite{sateesh2012metacognition,sateesh2016metacognition,Das2016AnIT,Anh2018WindSI}. Such neural networks have two components, cognitive and meta-cognitive. The latter component scrutinizes the former's knowledge and devises efficient learning strategies, namely the sample delete, neuron growth, parameter update, and sample reserve strategies to improve its learning ability. \cite{sateesh2012metacognition} implements the sample delete strategy for a classification problem as follows - when the predicted class label of the new training sample is the same as the actual class label, and the confidence level (estimated posterior probability) is greater than the expected value, the new training sample does not provide additional information to the classifier. It is deleted from the training pipeline without being used in the learning process. For a regression problem as in \cite{sateesh2016metacognition}, if the predicted error for the current sample is less than the delete threshold, then the knowledge content of the sample is similar to the knowledge present in the network. Thus, the sample is removed from the training sequence.

This paper has explored a sample selection strategy, which focuses on what object instances in each training image can be learned for object detection. The proposed method is based on the self-regulative learning process, which selects appropriate learning samples based on current knowledge, and the knowledge present in the training batch. The self-regulative learning process is evaluated using YOLO v3 Tiny architecture for object detection. The performance of the proposed meta-cognitive YOLO v3 Tiny is evaluated using the MS COCO dataset \cite{COCO} and compared against the performance of the baseline model. The results clearly highlight that meta-cognition helps achieve better performance without increasing the computational complexity of the model.  

The rest of the paper is organized as follows. Section \ref{sec_existingmethod} discusses our baseline model. Section \ref{sec_METAYOLO} describes the proposed method. Section \ref{sec_experiment} contains the experimental setup and performance evaluations. Finally, we conclude our paper in section \ref{sec_conclusion}.

\section{Background} 
\label{sec_existingmethod}
This section discusses the details of the baseline model. The architecture of YOLO v3 Tiny broadly consists of $13$ convolutional layers: feature extractor (seven layers) and feature detector (six layers). The backbone of YOLO v3 Tiny performs feature extraction, where each convolutional layer is followed by a batch normalization layer and Leaky ReLU activation. The feature extractor uses max-pooling layers to reduce the dimensionality of these convolutional layers. The network's backbone is followed by the detection head, which makes detections at two scales, i.e., output images of sizes $13 \times 13$ and $26 \times 26$ for an input image of resolution $416$. Each grid cell of an output image predicts three bounding boxes. The first output image of size $13 \times 13$, which detects large objects, is obtained at the tenth convolutional layer of the network. The feature map from the sixth pooling layer is upsampled and then concatenated with the feature map obtained at the fifth convolutional layer of the network. The concatenated feature map is then passed through the next two convolutional layers to give rise to the $26 \times 26$ image to detect medium sized objects. YOLO v3 Tiny performs detection based on these attributes, namely the class labels, object confidence score, and bounding box coordinates.

\section{Meta-cognitive YOLO V3 Tiny} \label{sec_METAYOLO}

In this section, the proposed method \textbf{M}eta-\textbf{C}ognitive \textbf{YOLO v3} \textbf{T}iny (MC-YOLOv3T) is discussed in detail. In MC-YOLOv3T, we explore a sample selection strategy of meta-cognitive learning in the training phase. Initially, we pre-process the training images by performing data augmentation and resizing. The pre-processed images are then passed to the network to obtain the predicted bounding boxes, following which we apply meta-cognitive thresholding to the class predictions of each bounding box. Meta-cognitive thresholding selectively samples the object instances in the training image. The learning threshold is an exponentially decaying function and is defined as follows:

\begin{equation}
	\label{EqChap3:thresholdfunction}
	N(t) = N_0 e^{-\lambda t},
\end{equation}
\noindent
where $N(t)$ is the threshold at epoch $t$, $N_0 = N(0)$ is the threshold at epoch $0$, and $\lambda$ is the decay constant. Since loss is higher in the initial phase of training, we set the meta-cognitive threshold to a higher value. As training proceeds and the loss decreases, we need to decrease the threshold value as well. Therefore, an exponentially decaying threshold function is used instead of a constant threshold.

The meta-cognitive threshold selectively samples the instances based on an error term. For every predicted bounding box assigned to a ground truth object, we calculate the error term which penalizes MC-YOLOv3T for assigning low scores to the correct classes and attaching high scores to the incorrect classes. The error term is defined as follows: 

\begin{equation}
	\label{EqChap3:errorfunction}
	E = \max (| p(0) - \widehat{p}(0)|, | p(1) - \widehat{p}(1)|, \hdots , | p(c) - \widehat{p}(c)|),
\end{equation}
\noindent
where the error term $E$ is the maximum of the absolute difference of the ground truth conditional class probability $p(0), p(1), \hdots, p(c)$ and predicted conditional class probability $\widehat{p}(0), \widehat{p}(1), \hdots,\widehat{p}(c)$, and $c$ is the maximum number of classes. If the error term is less than the meta-cognitive threshold, we set the classification loss contributed by the bounding box prediction to zero. This ensures that MC-YOLOv3T does not learn to classify an object instance if its prediction is within the error limit prescribed, thereby reducing overfitting. If the error term is more than the meta-cognitive threshold, we retain the classification loss contributed by such object instances. The loss function in MC-YOLOv3T comprises of three components - classification loss  \cite{YOLOv3} which penalizes the class predictions made by bounding boxes that predict objects, bounding box regression loss \cite{GIOU} which penalizes the model for the predicted bounding box coordinates, and objectness loss \cite{YOLOv3} which penalizes the model for the confidence with which it makes object predictions. These three losses are weighted as follows:

\begin{equation}
	\label{EqChap3:totalloss}
	\begin{aligned}
    \text{Total Loss} =  \alpha\text{ Classification loss} +  \beta\text{ Bounding box}\\ \text{ regression loss} + \gamma\text{ Objectness loss}
    	\end{aligned}
\end{equation}

Finally, we train the network by using the back-propagation algorithm. This training procedure is continued until the predefined number of epochs are reached, following which we use the trained network for detection. As the proposed method is based on YOLO v3 Tiny, its detection phase is identical to \cite{YOLOv3}. The complete training procedure is mentioned in Algorithm \ref{TrainAlgo}.

\begin{algorithm}
\caption{MC-YOLOv3T Training Procedure}\label{TrainAlgo}
\hspace*{\algorithmicindent} \textbf{Input:} Number of epochs $T$, Initial meta-cognitive threshold $N(0)$, Final meta-cognitive threshold $N(T)$, Hyper-parameters\\
\hspace*{\algorithmicindent} \textbf{Output:} Trained network 
\begin{algorithmic}[1]
\STATE $t \leftarrow 0$
\WHILE{$t \neq$ T}
\STATE Pre-process training images and pass through MC-YOLOv3T network
\STATE Obtain bounding box predictions
\STATE Compute meta-cognitive threshold, $N(t)$, by using Equation \ref{EqChap3:thresholdfunction} and error term, $E$, by using Equation \ref{EqChap3:errorfunction}
\STATE Compute classification loss for every bounding box assigned to a ground truth object
\STATE Compute regression loss for every bounding box assigned to a ground truth object
\STATE Compute objectness loss for every bounding box
\FOR{Bounding box predictions which are assigned to ground truth objects}
\IF{$E < N(t)$}
\STATE Classification loss $\leftarrow 0$
\ENDIF
\ENDFOR
\STATE $t \leftarrow t + 1$
\STATE Compute total loss by using Equation \ref{EqChap3:totalloss}
\ENDWHILE
\end{algorithmic}
\end{algorithm}
\noindent

% \begin{figure} [t]
% 	\begin{center}
% 		\includegraphics[width=0.6\linewidth]{Meta-Cognition-Flowchart}
% 		\caption{Flowchart of Meta-cognitive YOLO v3 Tiny}
% 		\label{fig:Meta-Cognition-Flowchart}
% 	\end{center}
% \end{figure}

\section{Experimental Results} \label{sec_experiment}
%\subsection{Experimental setup} \label{exp_setup}
We evaluate MC-YOLOv3T on MS COCO dataset \cite{COCO}. For evaluation, we use the metrics \cite{YOLOv3} of AP$_{50:95}$, AP$_{50}$, and AP$_{75}$, which denote the average precision at IoU thresholds of $0.5:0.95$, $0.5$ and $0.75$, respectively. We also use AP$_S$, AP$_M$ and AP$_L$, which denote the average precision for small, medium, and large objects, respectively. For pre-processing the images, we use the technique of mosaic data augmentation \cite{YOLOv4} and resize images for multi-scale training, where sizes are chosen from $\{320, 352, ..., 608\}$ in the multiples of $32$. Hyper-parameters of the network are listed in Table \ref{tab:HyperparameterValues}. Moreover, we use cosine scheduling for the learning rate and stochastic gradient descent optimizer. We also set the initial meta-cognitive threshold to $0.5$ and final threshold to $0.05$. First, MC-YOLOv3T is trained on the complete training set and $88\%$ of the validation set of MS COCO $2014$. We then perform validation on the remaining $12\%$ of the MS COCO $2014$ validation set. Further, the performance is evaluated using MS COCO $2017$ test set. The hardware acquired for training is an Intel Xeon $1.8$ GHz NVIDIA RTX Titan GPU with $128$ GB RAM and $500$ GB SSD for training the network.

\begin{table}[tbp]
	\scriptsize
	\centering
	\begin{tabular}{l|l}
		\hline
		Hyper-parameter & Values \\
		\hline
		Weight for regression loss ($\beta$)  & 3.54 \\
		Weight for classification loss ($\alpha$)  & 37.4 \\
		Weight for objectness loss ($\gamma$) & 64.3 \\
		Batch size & 32 \\
		Momentum & 0.937 \\ 
		Weight decay & 0.0005 \\
		Initial Learning Rate & 0.01 \\
		Final Learning Rate & 0.0005 \\
		Epochs & 600 \\
		\hline
	\end{tabular}
	\caption{Hyper-parameter values used for training}
	\label{tab:HyperparameterValues}
\end{table}

\subsection{Results} \label{results}
\begin{table*}[ht]
		\scriptsize
		\centering
		\begin{adjustbox}{width=1.0\textwidth,center}
		\begin{tabular}{l|llll|llll|llll}
			\hline
			& \multicolumn{4}{c|}{AP$_{50:95}$} & \multicolumn{4}{c|}{AP$_5$$_0$} & \multicolumn{4}{c}{AP$_7$$_5$} \\ \cline{2-13}
			& \makecell{Baseline\\model} & \makecell{MC-YOLOv3T\\300 epochs} & \makecell{MC-YOLOv3T\\600 epochs} & \makecell{Absolute\\delta} & \makecell{Baseline\\model} & \makecell{MC-YOLOv3T\\300 epochs} & \makecell{MC-YOLOv3T\\600 epochs} & \makecell{Absolute\\delta} & \makecell{Baseline.\\model} & \makecell{MC-YOLOv3T\\300 epochs} & \makecell{MC-YOLOv3T\\600 epochs} & \makecell{Absolute\\delta} \\
			\hline
			YOLO v3 Tiny@320 & 15.5 & 17.6 & \textbf{17.8} & \textbf{+2.3} & 29.5 & 30.5 & \textbf{31} & \textbf{+1.5} & 14.6 & 17.6 & \textbf{18.1} & \textbf{+3.5} \\
			YOLO v3 Tiny@416 & 17.5 & 20.2 & \textbf{20.4} & \textbf{+2.9} & 33 & 35 & \textbf{35.3} & \textbf{+2.3} & 17 & 20.2 & \textbf{20.8} & \textbf{+3.8} \\
			YOLO v3 Tiny@512 & 18.7 & 21.2 & \textbf{21.5} & \textbf{+2.8} & 35.3 & 37.3 & \textbf{37.9} & \textbf{+2.6} & 18 & 21.2 & \textbf{21.6} & \textbf{+3.6} \\
			YOLO v3 Tiny@608 & 19.4 & 21.5 & \textbf{21.7} & \textbf{+2.3} & 37 & 38.5 & \textbf{38.7} & \textbf{+1.7} & 18.6 & 21.2 & \textbf{21.5} & \textbf{+2.9} \\
			\hline
		\end{tabular}
		\end{adjustbox}
		\caption{Overview of improvement in AP achieved by meta-cognitive training, evaluated on MS COCO $2014$ validation set at multiple image resolutions}
		\label{tab:validation_results}
	\end{table*}

\begin{figure} [tbp]
	\begin{center}
		\includegraphics[width=1.0\linewidth]{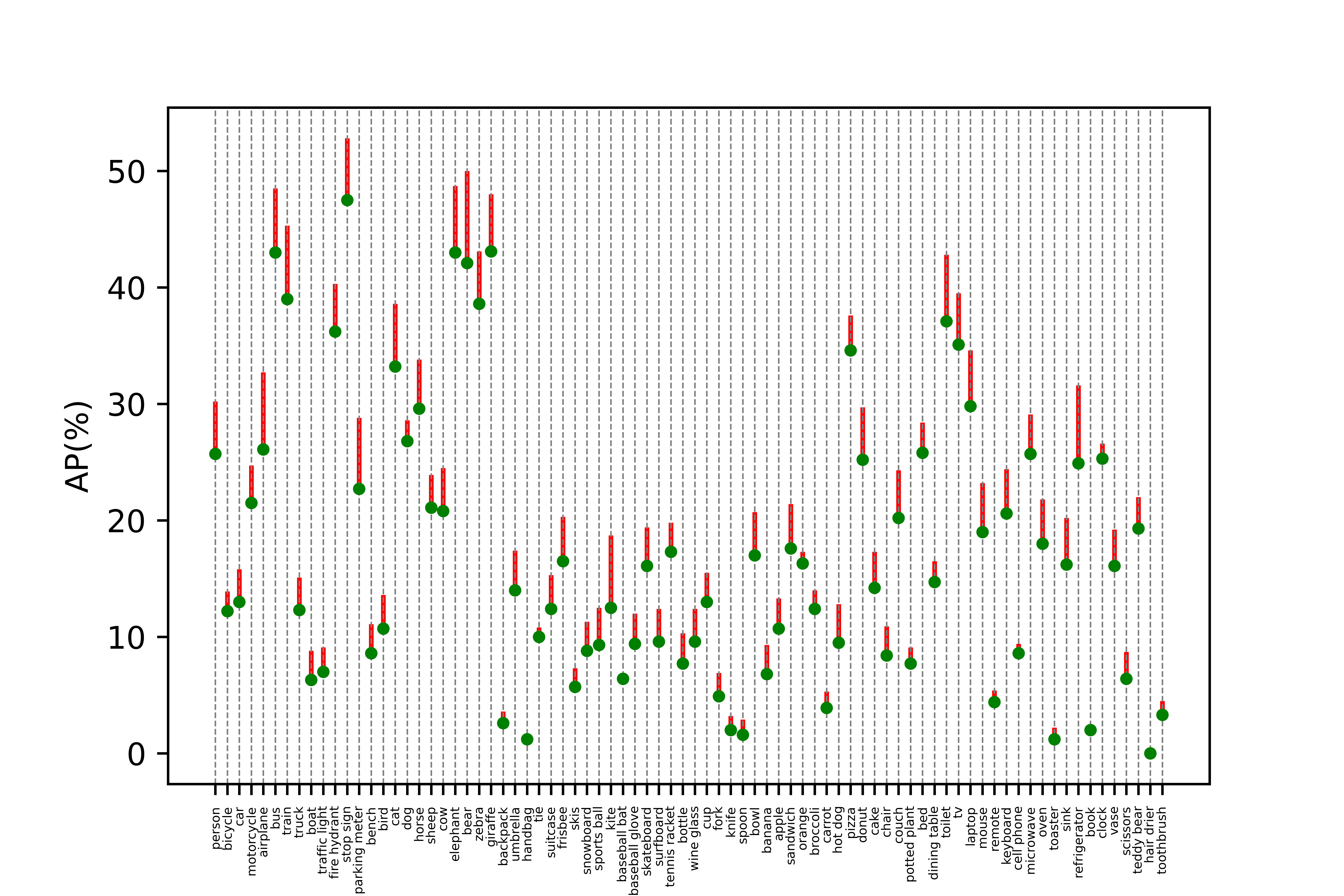}
		\caption{AP Improvement per class of COCO dataset. Green dots indicate performance of baseline model, red lines indicate performance gain using MC-YOLOv3T}
		\label{fig:AP_Improvement_Per_Class}
	\end{center}
\end{figure}

\begin{figure}
	\centering
	\begin{subfigure}{0.5\textwidth}
		\centering
		\includegraphics[width=0.95\linewidth,height=6cm]{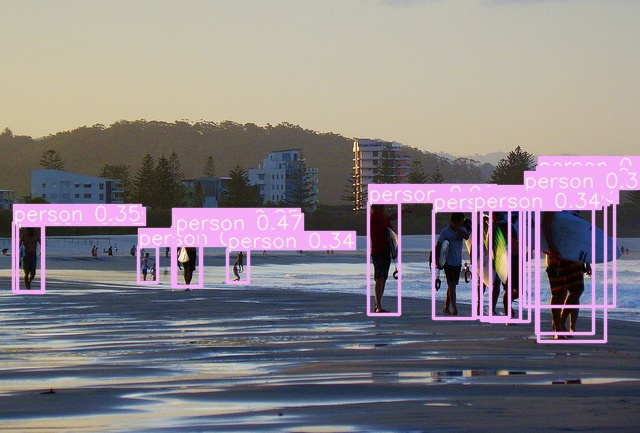}
		\caption{Detection results using the baseline model}
		\label{fig:regular_results}
	\end{subfigure}
	\begin{subfigure}{.5\textwidth}
		\centering
		\includegraphics[width=0.95\linewidth,height=6cm]{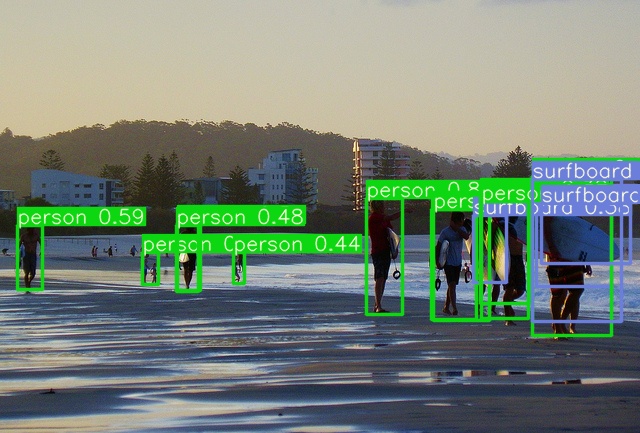}
		\caption{Detection results using MC-YOLOv3T}
		\label{fig:meta-cognition_results}
	\end{subfigure}
	\caption{Comparison of Detection Results as seen on a COCO 2017 Test set image}
	\label{fig:test}
\end{figure}

\begin{table}[tbp]
	\scriptsize
	\centering
	\begin{tabular}{l|lll}
		\hline
		& \makecell{Baseline\\model} & \makecell{MC-YOLOv3T\\600 epochs} & \makecell{Absolute\\delta} \\
		\hline
		AP$_{50:95}$ & 17.2 & \textbf{20.3} & \textbf{+3.1} \\
		AP$_5$$_0$ & 32.5 & \textbf{35.1} & \textbf{+2.6} \\
		AP$_7$$_5$ & 16.6 & \textbf{20.9} & \textbf{+4.4} \\
		AP$_S$ & 4.2 & \textbf{4.6} & \textbf{+0.4} \\ 		
		AP$_M$ & 16.6 & \textbf{20.3} & \textbf{+3.7} \\
		AP$_L$ & 29.9 & \textbf{34.4} & \textbf{+4.5} \\
		\hline
	\end{tabular}
	\caption{Overview of improvement in AP achieved by meta-cognitive training, evaluated on MS COCO $2017$ test set at image resolution of $416$}
	\label{tab:test_results}
\end{table}

In Table \ref{tab:validation_results}, we compare the performance of MC-YOLOv3T against the baseline model at input image resolutions of $320$, $416$, $512$, and $608$ to juxtapose at different scales. 
As MC-YOLOv3T has used a meta-cognition approach during training, it has been exposed to fewer object instances than the baseline model. We find that the loss of the baseline model saturates at $300$ epochs, whereas MC-YOLOv3T saturates at $600$ epochs. Although MC-YOLOv3T saturates at $600$ epochs, it has outperformed the baseline model at $300$ epochs only. The results evaluated on MS COCO $2017$ test set are presented in Table \ref{tab:test_results}. Here, it has been observed that MC-YOLOv3T significantly outperforms the baseline model. On the primary detection metric of AP$_{50:95}$, we see an improvement of $3.1\%$. Moreover, there is an improvement of $2.6\%$ (minimum) and $4.4\%$ (maximum) on the metric of AP$_{50}$ and AP$_{75}$, respectively. These results suggest that MC-YOLOv3T yields better improvements at higher IoU thresholds, i.e., AP$_{75}$. Further, we compute AP$_S$, AP$_M$, and AP$_L$. In Table \ref{tab:test_results}, there is a marginal increment of $0.4\%$ for small objects detection (AP$_S$) due to the absence of a detection layer for small objects in MC-YOLOv3T. Moreover, significant improvements of $3.7\%$ and $4.5\%$ have been observed for medium (AP$_M$), and large (AP$_L$) objects detection, respectively. 

The results in Table \ref{tab:test_results} are the average precision over $80$ classes of the dataset and do not fully reflect per class AP changes. Therefore, we plot per class AP improvements in Figure \ref{fig:AP_Improvement_Per_Class}. In this figure, it can be observed that MC-YOLOv3T has exhibited an improvement for all classes. The highest increment of $7.9\%$ is observed for the class 'bear'. However, there is a marginal improvement for a few classes. The marginal improvement for such classes is due to data imbalance, as these classes have fewer annotated instances in the training set, as compared to other classes.   

% MC-YOLOv3T yields significant improvement for object classes which have an already high AP score, reinforcing our notion that our approach reduces overfitting for classes on which the model has already trained well. For the few classes which show marginal improvement, we can attribute this to the fact that these classes show poor performance in the native model, implying that they require further training. 

We have also illustrated the detection results in Figure \ref{fig:regular_results} and \ref{fig:meta-cognition_results} for the baseline model and MC-YOLOv3T, respectively. The latter makes detections on the ‘surfboard’ category, which are absent in the former model. Furthermore, confidence scores for the ‘person’ category are higher in MC-YOLOv3T. These results indicate the superiority of our proposed method over baseline due to meta-cognitive principles.

\section{Conclusion}\label{sec_conclusion}
In this paper, we have employed a meta-cognitive learning-based sample selection strategy with YOLO v3 Tiny for object detection. The proposed MC-YOLOv3T, shows significant performance improvement across all input image resolutions. In particular, it yields greater improvements as the size of the object increases. Therefore, it has exhibited the best performance in case of large object detection. Finally, it is noteworthy that the MC-YOLOv3T achieves maximum 4.4\% improvement without any computational burden in the inference.
% \section{COPYRIGHT FORMS}
% \label{sec:copyright}

%\vfill\pagebreak

% \section{REFERENCES}
\label{sec:refs}

\bibliographystyle{IEEEbib}
\bibliography{strings,refs}

\end{document}